%% file: main.tex
\pdfoutput=1

\documentclass[11pt]{article}

\usepackage[]{emnlp2021}

\input{math_commands.tex}

\usepackage{times}
\usepackage{latexsym}

\usepackage[T1]{fontenc}

\usepackage[utf8]{inputenc}

\usepackage{hyperref}
\hypersetup{
    colorlinks=true,
    linkcolor=blue,
    urlcolor=blue,
    citecolor=blue,
    }
\usepackage[framemethod=TikZ]{mdframed}
\usepackage{natbib}
\usepackage{url}            
\usepackage{booktabs}       
\usepackage{amsfonts}       
\usepackage{nicefrac}       
\usepackage{microtype}      
\usepackage{xcolor}         
\usepackage{pgfplots}
\usepackage{amsmath,amsthm,amsfonts,amssymb,bm,stmaryrd}       
\usepackage{enumitem}
\usepackage[noend]{algpseudocode}
\usepackage{algorithm}
\usepackage{algorithmicx}
\usepackage{mathtools}
\usepackage{nccmath}
\usepackage{multirow}
\usepackage{bigdelim}
\usepackage{color, colortbl}
\usepackage{subcaption}
\renewcommand\cite{\citep}
\usepackage{longtable}
\usepackage{amsthm}
\theoremstyle{plain}

\newtheorem*{theorem*}{Theorem}
\newtheorem*{definition*}{Definition}
\newtheorem{question}{Question}
\newtheorem*{example*}{Example}

\algtext*{EndWhile}
\algtext*{EndIf}
\algtext*{EndFor}
\algtext*{EndFunction}
\usepackage{algorithm,algpseudocode}

\everypar{\looseness=-1}

\title{Towards Robust and Efficient Continual Language Learning}

\author{Adam Fisch$^{1,}$\thanks{$^*$Work done while an intern at Google DeepMind.}\quad Amal Rannen-Triki$^{2}$\quad Razvan Pascanu$^{2}$ \quad J\"org Bornschein$^{2}$ \\\\
\textbf{Angeliki Lazaridou$^{2}$\quad  Elena Gribovskaya$^{2}$\quad Marc'Aurelio Ranzato$^{2}$} \\\\
$^1$MIT CSAIL\qquad$^2$Google DeepMind}

\begin{document}

\maketitle

\input{sections/abstract}
\input{sections/intro}
\input{sections/related}
\input{sections/setting}

\input{sections/benchmark}

\input{sections/method}
\input{sections/results}

\input{sections/challenges}
\input{sections/conclusion}

\bibliography{anthology, custom}
\bibliographystyle{acl_natbib}

\appendix
\counterwithin{figure}{section}
\counterwithin{table}{section}
\counterwithin{algorithm}{section}
\input{sections/appendix/benchmark}
\input{sections/appendix/features}
\input{sections/appendix/additional_results}
\end{document}

%% file: math_commands.tex
\usepackage{amsmath,amsfonts,bm}

\def\eqref#1{equation~\ref{#1}}

\def\1{\bm{1}}

\DeclareMathAlphabet{\mathsfit}{\encodingdefault}{\sfdefault}{m}{sl}
\SetMathAlphabet{\mathsfit}{bold}{\encodingdefault}{\sfdefault}{bx}{n}



%% file: sections/abstract.tex
\begin{abstract}

As the application space of language models continues to evolve, a natural question to ask is how we can quickly adapt models to new tasks. We approach this classic question from a continual learning perspective, in which we aim to continue fine-tuning models trained on past tasks on new tasks, with the goal of ``transferring'' relevant knowledge. However, this strategy also runs the risk of doing more harm than good,  i.e., negative transfer. In this paper, we construct a new benchmark of task sequences that target different possible transfer scenarios one might face, such as a sequence of tasks with high potential of positive transfer, high potential for negative transfer, no expected effect, or a mixture of each. An ideal learner should be able to maximally exploit information from all tasks that have any potential for positive transfer, while also avoiding the negative effects of any distracting tasks that may confuse it.  We then propose a simple, yet effective, learner that satisfies many of our desiderata simply by leveraging a selective strategy for initializing new models from past task checkpoints. Still, limitations remain, and we hope this benchmark can help the community to further build and analyze such learners.\looseness=-1
\end{abstract}

%% file: sections/intro.tex
\section{Introduction}
\label{sec:intro}

Recent advances in large, pre-trained language models (LMs) have re-defined the ways practitioners approach and solve tasks in language understanding and generation~\cite[][\emph{etc}]{devlin-etal-2019-bert, raffel-etal-2020-t5, brown-etal-2020-gpt3, rae-etal-2021-gopher, hoffmann-etal-2022-chinchilla, chowdhery-etal-2022-palm}. Autoregressive language modeling removes the need for bespoke neural architectures, and provides a flexible framework for expressing diverse and complex tasks with unified input and output formats. At scale, LMs have achieved state-of-the-art performance across nearly every widely-used natural language processing (NLP) benchmark, and have had widespread impact in popular applications such as \texttt{ChatGPT}~\cite{chatgpt}. 

\begin{figure*}[!t]
    \centering
    \begin{mdframed}[roundcorner=10pt]
    \includegraphics[width=1\linewidth ]{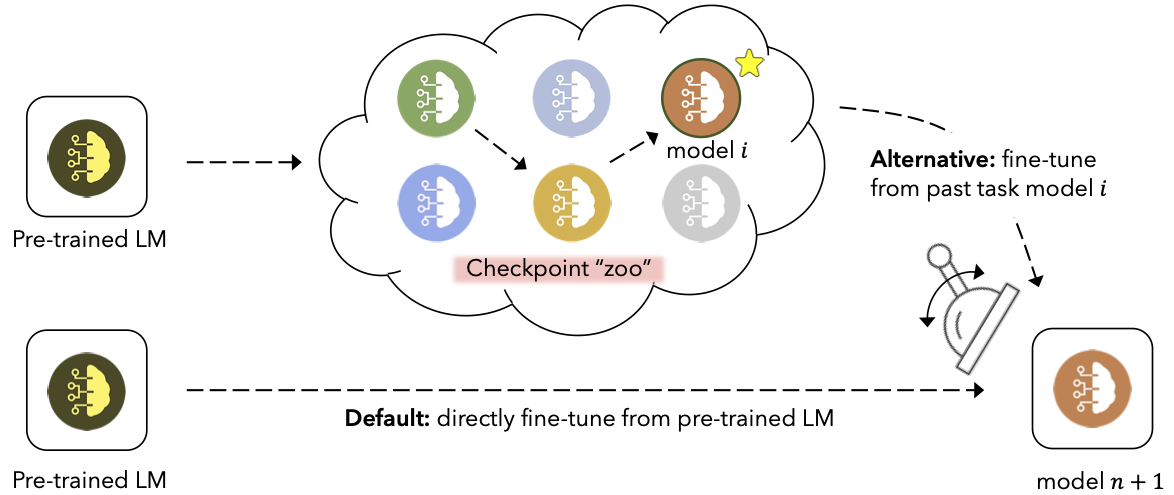}
    \end{mdframed}
    \caption{An illustration of our continual learning framework. When training the ($n+1$)th model we  choose between initializing from the default pre-trained language model and a previously fine-tuned model. This is repeated for each new task, and models within the zoo may build off each other to create a chain of fine-tuned models. Our motivation  is to make fine-tuning more efficient, while also being robust to the composition of previous tasks.\looseness=-1}
    \vspace{-10pt}
\end{figure*}

Though much interest has focused on the few-shot and zero-shot reasoning abilities of very large LMs, an effective approach to solving targeted NLP tasks is still to take the parameters of a pre-trained model, and fine-tune them on data from the new task~\cite{raffel-etal-2020-t5, gao-etal-2021-making, wei2022finetuned}. Here there is a similar trend: performance generally improves as the LM grows. Unfortunately this also results in significant computational costs during fine-tuning, even if the number of updates required is ultimately less than would be required if training the model from scratch. Furthermore, fine-tuning is typically performed independently for each new task, and ignores any other  tasks the LM might have previously been applied to. This not only leads to an accrual of computational cost over all  tasks, but also fails to \emph{share} acquired knowledge across tasks. 
In this work, we revisit fine-tuning efficiency from a continual learning perspective, motivated by the following question:\looseness=-1
\vspace{1pt}
\begin{question}
\label{q:ideal}
Suppose that we have already solved a set of $n$ previous tasks, $\{t_1, \ldots, t_n\}$.
Can we leverage any of the information gained from these tasks to solve the next task $t_{n+1}$ more efficiently?\looseness=-1
\end{question}
\vspace{1pt}
 Specifically, we study  the setting where each of the previous $n$ tasks is associated with a substantial number of training examples (e.g., several thousand). This setting is common, but not well addressed by few-shot  prompting. Our conjecture, which has also previously found empirical support in various related NLP  settings \cite[][\emph{inter alia}]{phang2019sentence, poth-etal-2021-pre, choshen2022start}, is that the standard pre-trained model---which wide-spread wisdom uses as the default initialization for any fine-tuning task---might not in-fact be the best checkpoint to use. Rather, models derived from one (or a combination) of the previous tasks might work even better as a starting point. This assumes that ``knowledge transfer'', or a form thereof, is accomplished via parameter initialization.\looseness=-1

We measure performance by how quickly our learning algorithm can produce good models for new tasks. Specifically, how much computational budget do we need to produce a model with some desired performance level? Or, alternatively, for a given computational budget, what is the best performance that we can achieve? Put in the context of past work on continual learning~\cite{continualworld, veniat2021efficient, nevis}, we are focused on forward transfer, which we define as having a faster ``rate of learning'' on a new task---relative to our baseline strategy of independent fine-tuning from the original pre-trained LM parameters. 
Naturally, how well one can hope to do depends  not only on the algorithm that is used, but also on the relationships between the new task and previous ones. We conduct a large-scale analysis of pairwise interactions across 55 popular and publicly available (English) language tasks using a T5 LM~\cite{raffel-etal-2020-t5}. Here we first fine-tune a T5 LM on task $\mathbf{A}$, and then proceed to fine-tune it on task $\mathbf{B}$. This indeed results in a fairly tumultuous transfer landscape: in some cases pre-training on $\mathbf{A}$ first can result in faster adaptation to task $\mathbf{B}$, but in other cases it can be quite detrimental. How can we expect which situation we may encounter, especially when faced with not just one, but many previous tasks and task combinations to transfer from?\looseness=-1

We argue that practical, efficient continual learning demands algorithms that are robust to the inevitable variations in the composition of the previous $n$ tasks. To this end, guided by our pairwise matrix of task interactions, we construct a challenging benchmark of multiple task sequences $(t_1, t_2, \ldots)$ that target different possible scenarios one might face, such as a sequence of tasks with high potential positive transfer, high potential for negative transfer, no expected effect, or a mixture of each. An ideal continual learner should be able to  exploit information from all tasks that have any potential for positive transfer, while also avoiding the harmful effects of any ``distractor'' tasks that may confuse it (and result in negative transfer).\looseness=-1

As a first step,  we propose a  simple method  that manages to satisfy many of our desiderata. Concretely, we  learn a {checkpoint selection} model that, given some representation of the current task $t_{n+1}$ and for all previously seen tasks $(t_1, \ldots, t_{n})$, predicts which previously saved checkpoint is the best checkpoint to initialize from---including the default option ``$t_0$'', which is simply the pre-trained model that a standard fine-tuning approach would start from. We demonstrate that training a lightweight gradient boosted decision tree~\cite{friedman2001gbdt} on top of (fast and easy to derive) features of each task over a small collection of held-out task pairs with different positive, negative, or neutral pairwise transfer relationships can result in good selection performance on new tasks: particularly when there exist  \emph{harmful} past tasks that are best to be ignored.\looseness=-1


In short, the core idea and contribution of this work can be summarized quite simply:
\begin{enumerate}[leftmargin=*, noitemsep]
\item We motivate and explore continual learning  for efficient LM fine-tuning and forward transfer;\looseness=-1\vspace{5pt}
\item To support this direction, we present a large-scale analysis of pairwise task transfer interactions, and a new benchmark of task sequences that capture diverse potential transfer profiles;\vspace{5pt}
\item Finally, we give a simple but effective method for checkpoint selection and model initialization that helps enable more robust forward transfer.\looseness=-1
\end{enumerate}

%% file: sections/related.tex
\section{Related work}
\label{sec:related}
\paragraph{Forward transfer.} This work builds on a large body of recent work that seeks to improve the efficiency of training modern language models through forward transfer (via parameter initialization). In particular,  leveraging auxiliary task data to improve target task performance has been a very active area of research years over the past few years ~\cite{luong-2016-mts, bingel-sogaard-2017-identifying,  phang2019sentence,wang-etal-2019-tell, gururangan-etal-2020-dont, pruksachatkun-etal-2020-intermediate,  vu-etal-2020-exploring, chang-lu-2021-rethinking-intermediate, aribandi2022ext}. Our work falls under the category of \emph{intermediate fine-tuning}, where a model is first trained on some auxiliary task $\mathbf{A}$ before being transferred to a target task $\mathbf{B}$. This paradigm has been well-analyzed in the pair-wise setting (i.e., $\mathbf{A} \rightarrow \mathbf{B}$ only), and multiple past studies have given empirical guidelines on how to select optimal transfer pairs~\cite[][]{ruder-plank-2017-learning, deshpande2021linearized,  poth-etal-2021-pre, huang2022frustratingly, choshen2022start, you_logme_2021, you_ranking_2022}. Here, we extend intermediate task training in a pair-wise fashion to training over full sequences of intermediate (and continually learned) tasks.\looseness=-1

\paragraph{Continual learning.} The key focus of this work is on continual learning for efficient language learning. Over the past decade, continual learning research has received significant interest within the wider machine learning community; see, e.g., \citet{parisi2019review} for a  review. Methodology-wise, existing work on continual learning can be approximately categorized into (a) replay-based~\cite{autume-2019-episodic, scialom-etal-2022-fine}, (b) regularization-based~\cite{kirkpatrick2017ewc, chaudhry2018efficient, qin-etal-2022-elle, ke2023continual}, or (c) architecture-based~\cite{carlson2010neverending, veniat2021efficient, douillard2021dytox, razdaibiedina2023progressive, qin2023recyclable} approaches. Many of these methods are motivated both by parameter-efficient forward transfer, as well as resistance to catastrophic forgetting. In contrast, similar to \citet{nevis}, we are only interested in \emph{training} efficiency on the new task---without worrying about how performance might suffer on previous tasks
---and focus only on different model initialization strategies for simplicity.\footnote{This is motivated by our simplifying assumptions that we (a) know what task we are trying to solve at each point in time, and (b) can checkpoint and load past models when necessary.} Our benchmark of transfer sequences also adds to a growing collection of continual learning datasets and analysis techniques for NLP~\cite{lazaridou2021mind, Livska2022StreamingQAAB, jang2022towards, wu2022pretrained}, with an attention towards sequences of a particular challenging structure that stress test for robustness to negative transfer.\looseness=-1


\paragraph{Efficient training for NLP.} Finally, our work is also  more broadly related to efficient training in language models~\cite{mattson2020mlperf, geiping2022cramming, menghani2023efficient}, which also includes efficiency in terms of parameter reuse and stored model size~\cite{Houlsby2019ParameterEfficientTL,li-liang-2021-prefix, he2022towards, hu2022lora, lei2023conditional}. While we do not consider these forms of efficiency in this work, they can be complementary to the form of training efficiency that we do concentrate on. Some of the metrics and analysis we propose may also be of independent interest.\looseness=-1

%% file: sections/setting.tex
\section{Problem formulation}

Let $\mathrm{LM}_{\theta} \colon \mathcal{X} \rightarrow \mathcal{Y}$ be our parametric language model, which generates natural language responses $y \in \mathcal{Y}$ given a prompt $x \in \mathcal{X}$. All of our experiments use the pre-trained T5 base model of~\citet{raffel-etal-2020-t5}, specifically the version adapted for language modeling by \citet{lester-etal-2021-power}.\looseness=-1

\subsection{Fine-tuning efficiency}
During fine-tuning, the model parameters vary as a function of the number of update steps $s\in \mathbb{N}$ that have been taken, denoted as $\theta(s)$. We quantify the time-dependent performance of our updating language model, $\mathrm{LM}_{\theta(s)}$, by its \textbf{best} (i.e., minimum) loss achieved within a  budget  of $B$ update steps:

\begin{align}
&\mathrm{Perf}(B) := \\
&\hspace{0.4cm}\min \Big \{ \underbrace{\mathbb{E}_{X,Y}\left[\ell(\mathrm{LM}_{\textcolor{black}{{\theta(s)}}}(X), Y)\right]}_{\textrm{average loss after step $s$}} \colon \textcolor{black}{s} \leq B \Big\}, \nonumber
\end{align}
where $\ell \colon \mathcal{Y} \times \mathcal{Y} \rightarrow \mathbb{R}$ is an arbitrary loss metric.

\input{sections/algorithm}

To reduce complexity and confounders between implementations, we choose to use the same model architecture, batch size, and learning rates in all of our experiments. The consequences of this are that (1) the number of update steps is directly proportional to the total training cost of the model, and (2) achieving better $\mathrm{Perf}(B)$ simply reduces to finding a better \emph{initialization} for our model, i.e., a choice of $\theta(0)$ that gives rise to efficient trajectories $\theta(s)$.\footnote{Note that an interesting direction for future work is to explore how the findings presented here generalize across different classes of models/learning algorithms (and if not, why).\looseness=-1}

Finally, as an aggregate measure of $\mathrm{Perf}(B)$ across budgets $B$, we evaluate the area under the  performance curve as a function of $\log$ updates, up to a maximum number of updates $B_\mathrm{max}$, i.e.,
\begin{align}
\mathrm{PerfAUC}(B_\mathrm{max}):= 
\int_{0}^{\log{B_{\mathrm{max}}}} \mathrm{Perf}(e^b)db.
\end{align}
$\mathrm{PerfAUC}(B_\mathrm{max})$ will be our primary metric for comparing methods, where we set $B_\mathrm{max} = 10k$, which is empirically the point for which  the majority of models for our tasks have (nearly) converged. More specifically, we will be interested in measuring the \emph{relative} efficiency of continual learning methods compared to the baseline of independent fine-tuning (where we always start from the same general pre-trained model for each task). Inspired by the relative forward transfer metrics of \citet{continualworld}, we compute this relative score as
\begin{align}
\label{eq:relative-perfauc}
    \frac{\mathrm{PerfAUC}(B_\mathrm{max})_{\mathrm{ind}} - \mathrm{PerfAUC}(B_\mathrm{max})_{\mathrm{m}}}{ \mathrm{PerfAUC}(B_\mathrm{max})_{\mathrm{ind}} - L},
\end{align}
where $(\cdot)_\mathrm{m}$, $(\cdot)_\mathrm{ind}$ are the  scores of the method and the baseline of independent fine-tuning, respectively, and $L$ is the metric lower bound for $\mathrm{PerfAUC}(B_\mathrm{max})$ (e.g., $0\% \times \log B_\mathrm{max}$ for error rate). Intuitively, this score measures the relative \emph{improvement} in terms of how much the compared method reduces the performance gap to the oracle (i.e., perfect predictions starting from step 0).  \looseness=-1

\subsection{Sequential fine-tuning}

As a starting point for the remainder of this paper, we now describe a very simple continual learning procedure for  sequential fine-tuning on a stream of tasks $(t_1, t_2, \ldots)$, see also  Algorithm~\ref{alg:train}. Beginning from an initial pre-trained language model $\mathrm{LM}_{\theta_0}$, we sequentially adapt models $\mathrm{LM}_{\theta_i}$ one after the other by using the model learned on some previous task $t_{j < i}$ to initialize the model used on task $t_{i}$. Note that here we write $\theta_i$ to index the task parameters, and will use $\theta_i(s)$ to denote the task parameters as a function of the number of updates. As described earlier, we use the same model architecture, batch size, and learning rate for each task. The only setting that changes is the initialization. The ``na\"ive'' implementation of sequential fine-tuning is to simply select the most recent checkpoint, $\theta_{i-1}$, see Algorithm~\ref{alg:vanilla}. Of course, this procedure is not necessarily optimal, since the model parameters learned for a task $\mathbf{A}$ may not be a good initialization for another task $\mathbf{B}$. In the next section we present an analysis of when $\mathbf{A}$ does have potential to transfer well as a good initialization for $\mathbf{B}$, and when it does not.\looseness=-1

\input{sections/algorithm_vanilla}

%% file: sections/algorithm.tex
\definecolor{darkgreen}{rgb}{0.31, 0.47, 0.26}
\begin{figure}[!t]
    \centering
        \vspace{-10pt}
    \begin{minipage}{1\linewidth}
\begin{algorithm}[H]
\caption{Sequential fine-tuning}
\label{alg:train}
\begin{algorithmic}[1]
\State{$\theta_0 \leftarrow \text{Pre-trained model}$}
\For{$t_i \in t_1, t_2, \ldots$}
    \State{\small\textcolor{darkgreen}{\# Initialize starting point from a previous model.}}
    \State{$\theta_{i}(0) \leftarrow \textsc{select}(\theta_0, \ldots, \theta_{i-1})$}
    \State{\small\textcolor{darkgreen}{\# Update current model using task data $(X_i, Y_i)$.}}
    \For{$s \in 1, 2, \ldots, B$}
        \State{$\theta_{i}(s) \leftarrow \textsc{update}(\mathrm{LM}_{\theta_i(s - 1)}, X_i, Y_i)$}
    \EndFor
    \State{\small\textcolor{darkgreen}{\# Keep the best model from the past $B$ steps.}}
    \State{$\theta_i \leftarrow \underset{s ~\leq ~B}{\arg\!\min} ~\mathbb{E}_{X_i, Y_i}[\ell(\mathrm{LM}_{\theta_i(s)}(X_i), Y_i)]$}
\EndFor
\end{algorithmic}
\end{algorithm}

\end{minipage}
\vspace{-10pt}
\end{figure}

%% file: sections/algorithm_vanilla.tex
\begin{figure}[!t]
    \centering
    \vspace{-10pt}
    \begin{minipage}{1\linewidth}
\begin{algorithm}[H]
\caption{``Na\"ive'' sequential fine-tuning}
\label{alg:vanilla}
\begin{algorithmic}[1]
\Function{$\textsc{select}$}{$\theta_0, \ldots, \theta_{i-1}$}
\State{\small\textcolor{darkgreen}{\# Return the most recently trained model.}}
\State{\Return{$\theta_{i-1}$}}
\EndFunction
\end{algorithmic}
\end{algorithm}
\end{minipage}
\vspace{-10pt}
\end{figure}

%% file: sections/benchmark.tex
\begin{figure*}[!t]
    \centering
    \includegraphics[width=1\linewidth]{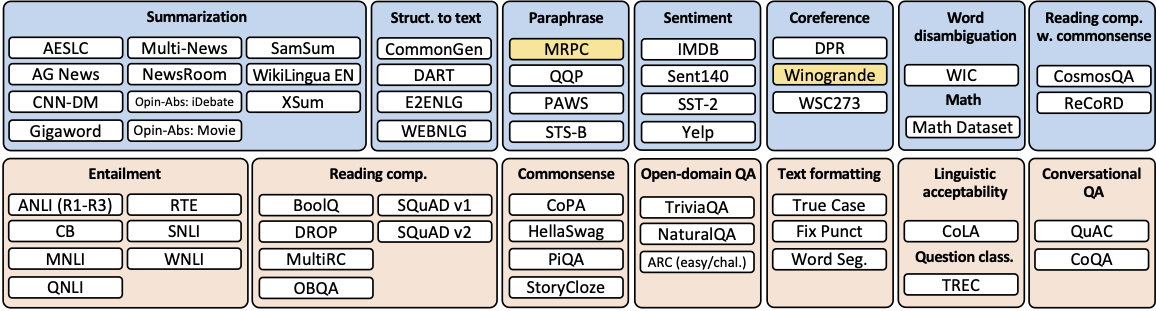}
    \caption{The collection of tasks used to create the sequential transfer benchmark used in this paper. Tasks are grouped into approximate ``families'', and families are further separated into training (top) and testing (bottom) splits. Highlighted training tasks are used for validation (i.e., task $\mathbf{C}$ when measuring transfer from $\mathbf{A} \rightarrow \mathbf{B} \rightarrow \mathbf{C}$).\looseness=-1}
    \label{fig:tasks}
\end{figure*}

\section{Analyzing task transfer potential}
\label{sec:transfer}

To help guide our understanding of how well  parameters learned from training on language task $\mathbf{A}$ perform when used as a starting point for training on a new language task $\mathbf{B}$, we conduct a large-scale analysis over various, diverse pairs of language tasks $(\mathbf{A}, \mathbf{B})$. Note that this is the \emph{minimal} setting for which sequential fine-tuning  can be applied.

\subsection{Dataset collection}

The tasks that we analyze are shown in Figure~\ref{fig:tasks}, and  mainly follow those used by FLAN~\cite{wei2022finetuned}, but without translation (we do not use multilingual models here). We use the same (loosely defined) ``task family'' groupings as \citet{wei2022finetuned} to help guide our analysis (below), but ultimately are interested in transfer between individual tasks.   To identify ``interesting'' pairs $(\mathbf{A}, \mathbf{B})$ that have either significantly negative or positive effects on each other, we use the following search strategy:\looseness=-1
\begin{enumerate}[leftmargin=*, noitemsep]
\item We evaluate {all} $16 \times 16$ task family pairs, where for a family pair $(\mathcal{F}_i, \mathcal{F}_j)$ we first train a model on a mixture of all tasks $t_i' \in \mathcal{F}_i$, and then use that model as the starting point for training the second model on a mixture of all tasks $t_j' \in \mathcal{F}_j$. Each  model is trained on a balanced mixture of task data, and evaluated according to the average performance across tasks within each family.\looseness=-1\vspace{5pt}

\item For a pair $(\mathcal{F}_i, \mathcal{F}_j)$ the average performance after sequentially fine-tuning on  $\mathcal{F}_j\rightarrow\mathcal{F}_j$ can either be better, worse, or approximately the same relative to training independently on $\mathcal{F}_j$. We use this signal as evidence that there may exist individual tasks $t_i', t_j' \in \mathcal{F}_i \times \mathcal{F}_j$ with a similar trend.\vspace{5pt}

\item For each family $\mathcal{F}_i$, we identify the top-$K$  families $\mathcal{F}_j$ with the \emph{best} average transfer to $\mathcal{F}_i$, as well as the worst-$K$ families $\mathcal{F}_j$ with the \emph{worst} average transfer to $\mathcal{F}_i$. $K$ is set to $3$. We then evaluate all individual task pairs in $\mathcal{F}_i \times \mathcal{F}_j \times \mathcal{F}_k$.
\end{enumerate}

\begin{figure}[!t]
    \centering
    \includegraphics[width=1\linewidth]{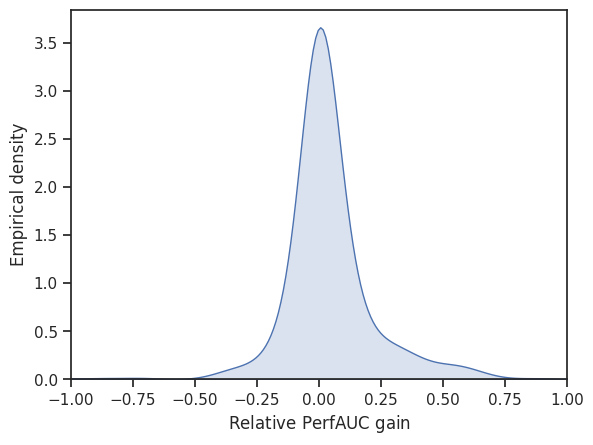}
    \caption{A density plot of the empirical distribution of the relative $\mathrm{PerfAUC}$ across the 1757 task pairs $\mathbf{A}\rightarrow\mathbf{B}$ that we (selectively) evaluate. All models are trained with ``na\"ive'' sequential fine-tuning, where we use the checkpoint of task $\mathbf{A}$ as a starting point for task $\mathbf{B}$.\looseness=-1}
    \label{fig:transfer_distribution}
\end{figure}

In total, we evaluate $1757$ unique task pairs. Figure~\ref{fig:transfer_distribution} plots the distribution of transfer results  in terms of relative $\mathrm{PerfAUC}$. Consistent with observations in prior work~\cite{pruksachatkun-etal-2020-intermediate, poth-etal-2021-pre},  we can see that while on many tasks there is no marked effect due to sequential fine-tuning, there do exist a significant tails of both positive and negative transfer instances.\footnote{Note, however, that this distribution is also artificially \emph{biased} towards the tails, due to our search and evaluation strategy. Nevertheless, it can still be inferred that substantial absolute numbers of both positive and negative instances exist.}\looseness=-1

\subsection{Types of transfer profiles}

\begin{figure*}[!t]
    \centering
    \begin{subfigure}[b]{0.32\textwidth}
    \includegraphics[width=1.0\linewidth]{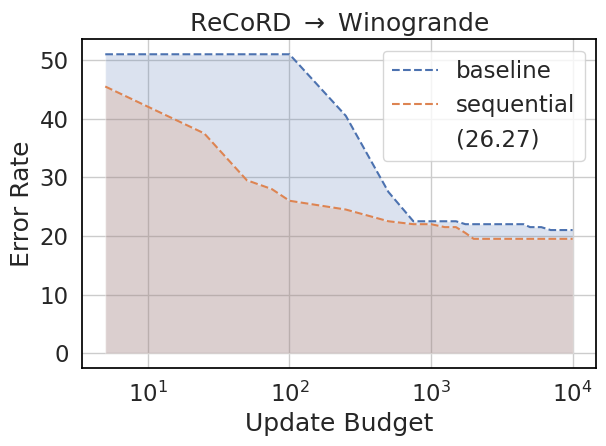} 
    \end{subfigure}
    \begin{subfigure}[b]{0.32\textwidth}
    \includegraphics[width=1.0\linewidth]{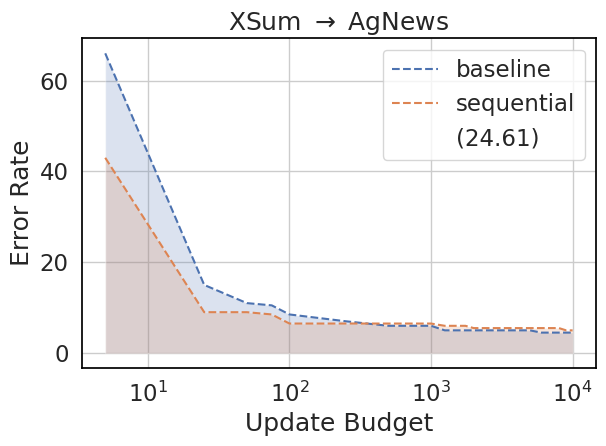}
    \end{subfigure}
    \begin{subfigure}[b]{0.32\textwidth}
    \includegraphics[width=1.0\linewidth]{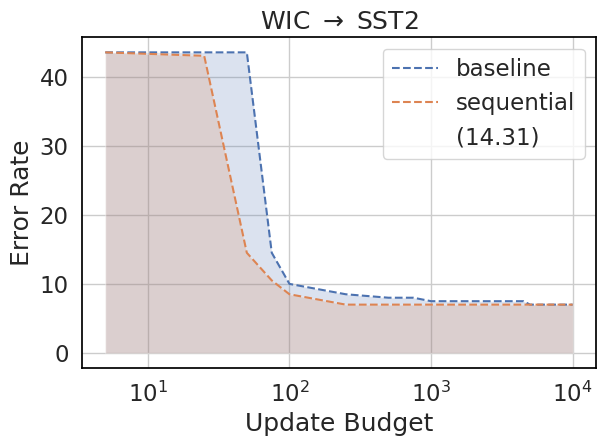} 
    \end{subfigure}
    
    \begin{subfigure}[b]{0.32\textwidth}
    \includegraphics[width=1.0\linewidth]{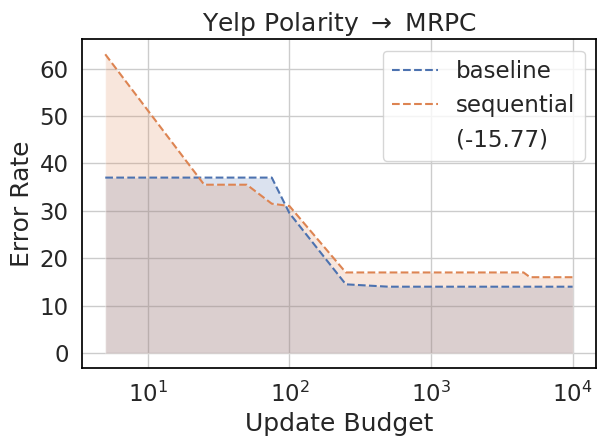} 
    \end{subfigure}
    \begin{subfigure}[b]{0.32\textwidth}
    \includegraphics[width=1.0\linewidth]{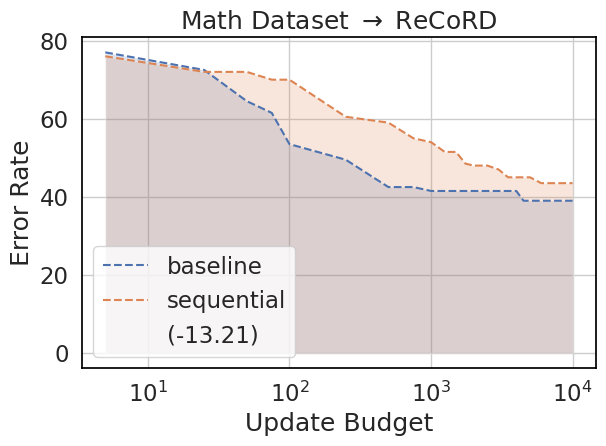}
    \end{subfigure}
    \begin{subfigure}[b]{0.32\textwidth}
    \includegraphics[width=1.0\linewidth]{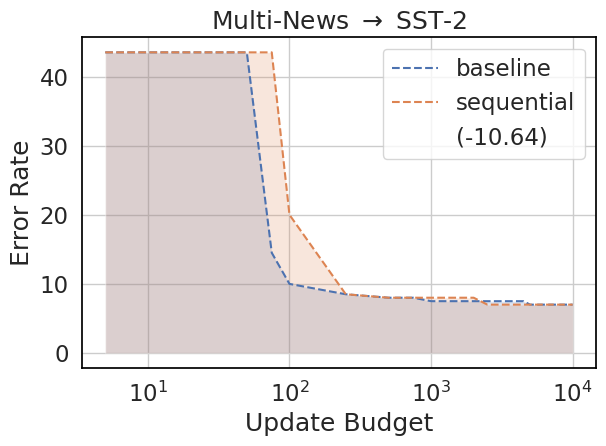} 
    \end{subfigure}
    \caption[]{Example positive and negative pairwise transfer profiles $\mathbf{A} \rightarrow \mathbf{B}$, in which the lowest loss per update budget on task $\mathbf{B}$ is plotted. Blue is for the baseline of independent fine-tuning (pre-trained model $\rightarrow \mathbf{B}$), while orange is for ``na\"ive'' sequential fine-tuning (pre-trained model $\rightarrow \mathbf{A} \rightarrow \mathbf{B}$). Relative $\mathrm{PerfAUC}$ is included in each legend.\looseness=-1}
     \label{fig:transfer_examples}
\end{figure*}


Figure~\ref{fig:transfer_examples} gives a number of qualitative examples that exhibit some of the different types of transfer profiles that arise. Some models exhibit strong transfer from the beginning: the 0-shot performance is good, and continues to improve. Good 0-shot performance, however, is not always a reliable indicator of future success: some models start out with better performance, but improve more slowly. Other pairs yield no tangible difference. There is also significant variation across tasks, with some tasks acting as fairly  ``universal donors'' with positive transfer to most other tasks, while others mostly result in negative, or at best minimal positive, transfer. For example, in our experiments,  $73\%$ of models trained first on the STS-B dataset~\cite{cer2017semeval} had $>{+}5\%$ relative $\mathrm{PerfAUC}$ across evaluated target tasks. On the other hand, $70\%$  of models trained first on the Math dataset~\cite{saxton2018analysing} had $< {-}5\%$ relative $\mathrm{PerfAUC}$ across evaluated target tasks.\looseness=-1

Interpreting and understanding \emph{why} these transfer curves form the way they do is tricky---and something we leave as an open problem for future research. Nevertheless, the existence of these empirical phenomena allows us to construct challenging sequences of tasks over which to perform continual learning. As previously discussed, an ideal learner should be able to exploit information from all tasks that have any potential for positive transfer (demonstrated by having a {positive pairwise} transfer result), while also avoiding the negative effects of any potentially harmful tasks that may confuse it (demonstrated by having a  {negative pairwise} transfer result). An ideal learner should be agnostic to the mechanism that is responsible for the positive or negative transfer, which in many common situations (such as in many of the tasks presented here) may not be that well understood.\looseness=-1

\begin{table}[!t]
\small
\centering
\resizebox{\linewidth}{!}{%
\begin{tabular}{@{}l|l|l@{}}
\toprule
$\mathbf{A} \rightarrow \mathbf{C}$ & $\mathbf{B} \rightarrow \mathbf{C}$ & Desired $\mathbf{A} \rightarrow \mathbf{B} \rightarrow \mathbf{C}$ \\ \midrule
Positive &
  Positive &
  $\geq \max (\mathbf{A} \rightarrow \mathbf{C}, \mathbf{B} \rightarrow \mathbf{C})$ \\
Positive & Negative & $\approx \mathbf{A} \rightarrow \mathbf{C}$  \\
Positive & Neutral  & $\approx \mathbf{A} \rightarrow \mathbf{C}$  \\\midrule
Negative & Positive & $\approx \mathbf{B} \rightarrow \mathbf{C}$  \\
Negative & Negative & $\approx \mathbf{C}$                         \\
Negative & Neutral  & $\approx \mathbf{C}$                         \\\midrule
Neutral  & Positive & $\approx  \mathbf{B} \rightarrow \mathbf{C}$ \\
Neutral  & Negative & $\approx \mathbf{C}$                         \\ \bottomrule
\end{tabular}%
}
\caption{Types of task triplets in our benchmark. 
The final column indicates the desired behavior on task $\mathbf{C}$ when using an ``ideal'' continual learning algorithm.\looseness=-1}
\label{tab:benchmark_format}
\vspace{-15pt}
\end{table}

\subsection{Constructing a diagnostic benchmark}

We leverage the pairwise transfer results to construct a series of diverse, diagnostic task sequences. The format of these sequences is outlined in Table~\ref{tab:benchmark_format}. We split the pairs of tasks into \textbf{positive}, \textbf{negative}, and \textbf{neutral} subsets based on the magnitude of their relative $\mathrm{PerfAUC}$ to the independent fine-tuning baseline (see Eq.~\ref{eq:relative-perfauc}). For positive/negative tasks, we attempt to account for variance in training (where the randomness is over the batch selection and ordering during SGD) by requiring that the mininum/maximum relative $\mathrm{PerfAUC}$ results across all random trials are above/below ${+}5\%/{-}5\%$, respectively (though occasional false positives  exist across different runs). We then construct 8 different types of triplets $(\mathbf{A}, \mathbf{B}, \mathbf{C})$, where each of the preceding tasks $\mathbf{A}$ and $\mathbf{B}$ are mostly positive, negative, or neutral \emph{pairwise} transfer sources for the target task $\mathbf{C}$ (i.e., $\mathbf{A} \rightarrow \mathbf{C}$ and $\mathbf{B}\rightarrow \mathbf{C}$, respectively). Note that we exclude the neutral/neutral case. For each  configurations, we include multiple sets of triplets with different source and target tasks, and measure the median performance across task instances in all experiments. Specifically, on the test split of the benchmark, for each type of triplet (e.g., positive/positive) we include $4$ distinct target tasks $\mathbf{C}$, each with $4$ distinct preceding task pairs $(\mathbf{A}, \mathbf{B})$, for a total of $16$ triplets per setting (and $128$ triplets in total). Additional details are provided in Appendix~\ref{app:benchmark}.\looseness=-1

%% file: sections/method.tex
\section{Learning a checkpoint selector}

We now propose a straightforward, but effective, algorithm for robust forward transfer. Motivated by our analysis in Section~\ref{sec:transfer}, we consider a  simplified version of Question~\ref{q:ideal} that we posed in Section~\ref{sec:intro}:

\vspace{2pt}
\begin{question}
Suppose that the set of previously solved tasks $\{t_1, \ldots, t_n\}$ contains a distinct set of tasks with trained models $\theta_i$ that act as good initializations for $t_{n+1}$, i.e. $\mathcal{P} \subseteq \{t_1, \ldots, t_n\}$. Given features $\phi(t_i, t_{n+1}) \in \mathbb{R}^d$, can we learn a discriminator $\mathcal{D} \colon \mathbb{R}^d \rightarrow \{0, 1\}$ to identify ``positive'' task candidates $t_i \in \mathcal{P}$ to leverage for learning $t_{n+1}$?
\end{question}
\vspace{2pt}

 Concretely, when training a new model for $t_{n+1}$, we seek to allow ourselves to select a previously fine-tuned model on some task $t_i \in \{t_1, \ldots, t_n\}$ to initialize from---if we think that it will lead to \emph{positive} transfer. If multiple such tasks exist, then we select the most confident one, using confidence scores from some model $\mathcal{C} \colon \mathbb{R}^d \rightarrow [0, 1]$, which is typically the same underlying model as $\mathcal{D}$, but without a decision threshold. If we are not confident that any such task model exists, then we initialize from the default pre-trained language model. This process, which we call \emph{selective} sequential fine-tuning, is illustrated in Algorithm~\ref{alg:selective}, and is similar in spirit to prior work on checkpoint selection for transfer learning (see \S\ref{sec:related}), with the caveat that we only select candidates that pass a decision threshold (see also \S\ref{sec:challenges} for a discussion on the potential importance of properly calibrating this threshold). This process is repeated for each new task, e.g., for a sequence $\mathbf{A} \rightarrow \mathbf{B} \rightarrow \mathbf{C}$, task $\mathbf{A}$ is initialized from the pre-trained model, task $\mathbf{B}$ is either initialized from the pre-trained model or the checkpoint for $\mathbf{A}$, and task $\mathbf{C}$ is either initialized from the pre-trained model or either of the checkpoints for $\mathbf{A}$ or $\mathbf{B}$. In general, there are $2^n$ possible paths (in terms of sequential initializations) to take from $t_0$ to task $t_{n+1}$.\looseness=-1

We choose to instantiate $\mathcal{D}$ as a simple gradient boosted decision tree (GBDT)~\cite{friedman2001gbdt} operating on several light-weight ``meta'' features, $\phi(t_i, t_j)$ of an input task pair. $\mathcal{C}$ is the pre-binarized decision function of the GBDT.  The GBDT is trained over positive and negative pairs from the training split of our benchmark.\footnote{While we ignore this aspect in this work, note that this introduces a distributional shift at test time, since candidate models are themselves products of multiple iterations of this selection algorithm, rather than only pairwise transfer instances.}  The features $\phi$ are fairly conventional (e.g., similar motivation can be found in the related approaches of ~\citet{bingel-sogaard-2017-identifying, poth-etal-2021-pre}). They  include metadata (e.g., if any of the previous tasks are in the same family as $t_{j}$, the zero-shot and few-shot performance of model $t_i$ on $t_{j}$, and a number of gradient-based similarity metrics comparing updates to $t_i$ and $t_j$ relative to a $t_0$ starting point. See Appendix~\ref{app:features} for more details. We binarize $\mathcal{D}$ by thresholding the GBDT confidence at $0.5$ (i.e., we only consider a checkpoint to be a candidate for selection if it is judged by our model to be more likely than not to be a positive transfer pair).\looseness=-1

\input{sections/algorithm_seq}

%% file: sections/algorithm_seq.tex
\begin{figure}[!t]
    \centering
\vspace{-11.5pt}
\begin{minipage}{1\linewidth}
\begin{algorithm}[H]
\begin{algorithmic}[1]
\Function{$\textsc{select}$}{$\theta_0, \ldots, \theta_{i-1}$}
    \State{\small\textcolor{darkgreen}{\# Estimate selection of positive transfer candidates}}
    \State{\small\textcolor{darkgreen}{\# from the corresponding tasks $(t_1, \ldots, t_{i-1}. t_i)$, }}
    \State{\small\textcolor{darkgreen}{\# $\mathcal{D}$ is a trained ``positive transfer'' discriminator.}}
    \State{$\widehat{\mathcal{P}} \leftarrow \big\{t_j : \mathcal{D}(t_j, t_i) = 1, j < i\big\}$}

    \If{$\widehat{\mathcal{P}} \neq \varnothing$}
        \State{\small\textcolor{darkgreen}{\# Pick the most \underline{confident} candidate if any}}
\State{\small\textcolor{darkgreen}{\# exists, where $\mathcal{C}$ is a  confidence measure.}}
        \State{$j^* \leftarrow {\arg\!\max}_j\big\{\mathcal{C}(t_j, t_i) : t_j \in \widehat{\mathcal{P}}\big\}$}

    \Else
        \State{\small\textcolor{darkgreen}{\# Otherwise, default to the pre-trained model.}}
        \State{$j^* \leftarrow 0$}
    \EndIf
    \State{\Return{$\theta_{j^*}$}}
\EndFunction
\end{algorithmic}
\caption{``Selective'' sequential fine-tuning}
\label{alg:selective}
\end{algorithm}
\end{minipage}
\vspace{-10pt}
\end{figure}

%% file: sections/results.tex
\section{Results}

\setlength{\tabcolsep}{20pt} 
\begin{table*}[]
\centering
\begin{tabular}{l|l|lll}
\toprule
$\mathbf{A} \rightarrow \mathbf{C}$ & $\mathbf{B} \rightarrow \mathbf{C}$ & \multicolumn{3}{c}{$\mathbf{A} \rightarrow \mathbf{B} \rightarrow \mathbf{C}$} \\
Pairwise                        & Pairwise                        & Na\"ive & Selective & Oracle \\ \midrule
\cellcolor[HTML]{84D36B}34.5 & \cellcolor[HTML]{84D36B}31.8  & 42.7  & 42.4    & 43.6 \\
\cellcolor[HTML]{84D36B}13.9 & \cellcolor[HTML]{FFCCC9}-17.0 & -2.35  & 11.8     & 13.9 \\
\cellcolor[HTML]{84D36B}15.1  & \cellcolor[HTML]{EFEFEF}0.244  & 16.9  & 15.1     & 19.9 \\ \midrule
\cellcolor[HTML]{FFCCC9}-17.0 & \cellcolor[HTML]{84D36B}13.7  & 12.0  & 11.8     & 15.5 \\
\cellcolor[HTML]{FFCCC9}-24.5 & \cellcolor[HTML]{FFCCC9}-19.5 & -24.7 & 0.00     & 0.00  \\
\cellcolor[HTML]{FFCCC9}-17.0 & \cellcolor[HTML]{EFEFEF}-1.65  & -8.34  & 0.00     & 0.00  \\ \midrule
\cellcolor[HTML]{EFEFEF}-0.703  & \cellcolor[HTML]{84D36B}25.2  & 19.8  & 15.0     & 26.3 \\
\cellcolor[HTML]{EFEFEF}-1.88  & \cellcolor[HTML]{FFCCC9}-16.9 & -12.6 & 0.00     & 0.00  \\ \bottomrule
\end{tabular}%
\caption{$\mathrm{PerfAUC}$ results on our benchmark sequences. Each row is the median of all 16 instances of that configuration (e.g., positive $\mathbf{A} \rightarrow$ positive $\mathbf{B} \rightarrow \mathbf{C}$. Green denotes intended  ``positive'' pairwise transfer, red denotes ``negative'' pairwise transfer, while grey denotes ``neutral'' transfer (i.e., no substantial effect). Oracle is the best achievable result using any (possible) sequence of checkpoints from the initial pre-trained model to task $\mathbf{C}$. A score of 0 means performing as well as a model that fine-tunes from the original pre-trained model, while positive/negative scores are improvements/degradations relative to that, which is the default used today.\looseness=-1}
\label{tab:results}
\vspace{-5pt}
\end{table*}

We compare the behavior of na\"ive sequential fine-tuning, our selective sequential fine-tuning procedure, and an oracle checkpoint selection algorithm across the  task sequences in our benchmark. Our results are given in Table~\ref{tab:results}. See also Appendix~\ref{app:additional_results}. The oracle picks the best sequential fine-tuning path from $t_0$ to $t_{n+1}$  in hindsight, and is used as an upper-bound to our selective model performance (as $n = 2$ in our experiments, this results in $4$ total possible paths). We report the median relative $\mathrm{PerfAUC}$ result across all 16 triplets for each sequence type (e.g., for the triplet  $\mathbf{A} \rightarrow \mathbf{B} \rightarrow \mathbf{C}$, where $\mathbf{A} \rightarrow \mathbf{B}$ results in positive pairwise transfer, while  $\mathbf{B} \rightarrow \mathbf{C}$ results in negative pairwise transfer).\looseness=-1

\paragraph{Forward transfer.} Rows in Table~\ref{tab:results} with green entries denote sequence types with potential for positive forward transfer. When both tasks $\mathbf{A}$ and $\mathbf{B}$ are positive intermediate tasks for task $\mathbf{C}$, continuing to fine-tune $\mathbf{A} \rightarrow \mathbf{B} \rightarrow \mathrm{C}$ generally also result in positive transfer---interestingly, often to a larger degree than either of only $\mathbf{A} \rightarrow \mathrm{C}$ or $\mathbf{B} \rightarrow \mathrm{C}$. When a positive intermediate task is paired with a \emph{negative} intermediate task (red entries), the performance of na\"ive sequential fine-tuning is sensitive to their ordering (and is better when the most recently trained task is a positive transfer pair). Our selective procedure, however, manages to leverage positive transfer where possible regardless of order---though it can significantly lag behind the oracle in certain cases.\looseness-1

\paragraph{Negative transfer.} Rows in Table~\ref{tab:results} with red entries denote sequence types with potential for  negative transfer. Fortunately, unlike sequential transfer with positive transfer options, harmful effects from two negative, or negative and neutral, intermediate tasks $\mathbf{A}$ and $\mathbf{B}$  rarely compound (in fact, the negative effect can sometimes be attenuated). In the cases where there are no positive intermediate tasks to transfer from, our selective algorithm is successful in choosing the pre-trained model as a starting checkpoint (resulting in $0$, but at least not \emph{negative}, relative $\mathrm{PerfAUC}$).\looseness=-1

%% file: sections/challenges.tex
\section{Limitations and challenges}
\label{sec:challenges}
While our work provides a starting point for testing robust and efficient continual learning, several limitations remain. Most significantly, our focus is restricted to T5 base models with simple optimization routines, and the only method of transfer that we test and explore is via parameter initialization, without considering space efficiency (i.e., in reducing the number of saved parameters across all tasks). Our selective checkpoint initialization strategy is therefore advantaged with respect to this particular setting. Additionally, our oracle is only evaluated for this strategy---other methods that use different knowledge transfer paradigms may do even better~\cite{ermis2022memory,qin2022lfpt, razdaibiedina2023progressive}. We note that noise is also introduced though stochastic effects in SGD (e.g, learning rates, batch sizes), which introduces some confounding effects, especially when integrating over $\log$ updates (which biases $\mathrm{PerfAUC}$ towards early performance). This is more significant for some tasks than others. Finally, another challenge is that as the number of considered tasks grows, our selective classifier may become more prone to identifying a large number of false positives. Without using calibration techniques that account for multiple testing~\cite[e.g.,][]{fisch2021efficient}, the selective classifier may choose poor checkpoints with increasingly high probability.\looseness=-1

%% file: sections/conclusion.tex
\section{Conclusion}

This paper develops a collection of task sequences with diverse transfer scenarios to test for efficient and robust continual learning on language tasks. Our benchmark targets different possible scenarios one might face: such as a sequence of tasks with high potential for positive transfer, negative transfer, no effect, or a mixture of each. As a first step, we proposed a selective algorithm for choosing past checkpoints to initialize from when considering each new task $t_{n+1}$. Limitations remain, and we hope this benchmark may help analyze and identify strong continual language learning algorithms.\looseness=-1

%% file: sections/appendix/benchmark.tex
\section{Benchmark details}
\label{app:benchmark}

\begin{table*}[]
\centering
\small
\begin{tabular}{@{}lllll@{}}
\toprule
Name                          & Train & Validation & Test   & Metric     \\ \midrule
anli                          & 76946 & 600        & 3200   & accuracy   \\
squad                         & 60000 & 400        & 22443  & f1         \\
mnli                          & 60000 & 400        & 19647  & accuracy   \\
hellaswag                     & 30000 & 200        & 10042  & accuracy   \\
yelp\_polarity\_reviews       & 30000 & 200        & 38000  & accuracy   \\
record                        & 30000 & 200        & 10000  & accuracy   \\
true\_case                    & 30000 & 200        & 3000   & mean\_edit \\
xsum                          & 30000 & 200        & 11301  & rougeLsum  \\
newsroom                      & 30000 & 200        & 108862 & rougeLsum  \\
multi\_news                   & 30000 & 200        & 5622   & rougeLsum  \\
web\_nlg\_en                  & 30000 & 200        & 1667   & rougeLsum  \\
qnli                          & 30000 & 200        & 5463   & accuracy   \\
winogrande                    & 30000 & 200        & 1267   & accuracy   \\
sentiment140                  & 30000 & 200        & 498    & accuracy   \\
dart                          & 30000 & 200        & 2768   & rougeLsum  \\
wiki\_lingua\_english\_en     & 30000 & 200        & 28614  & rougeLsum  \\
common\_gen                   & 30000 & 200        & 993    & rougeLsum  \\
word\_segment                 & 30000 & 200        & 3000   & mean\_edit \\
cnn\_dailymail                & 30000 & 200        & 11490  & rougeLsum  \\
e2e\_nlg                      & 30000 & 200        & 4299   & rougeLsum  \\
glue\_qqp                     & 30000 & 200        & 40430  & accuracy   \\
paws\_wiki                    & 30000 & 200        & 8000   & accuracy   \\
sst2                          & 30000 & 200        & 872    & accuracy   \\
snli                          & 30000 & 200        & 10000  & accuracy   \\
trivia\_qa                    & 30000 & 200        & 11313  & f1         \\
ag\_news\_subset              & 30000 & 200        & 7600   & accuracy   \\
quac                          & 30000 & 200        & 7354   & nll        \\
math\_dataset                 & 30000 & 200        & 10000  & accuracy   \\
drop                          & 30000 & 200        & 9536   & f1         \\
fix\_punct                    & 30000 & 200        & 3000   & mean\_edit \\
natural\_questions            & 30000 & 200        & 3610   & f1         \\
gigaword                      & 30000 & 200        & 1951   & rougeLsum  \\
multirc                       & 27043 & 200        & 4848   & accuracy   \\
cosmos\_qa                    & 25062 & 200        & 2985   & accuracy   \\
imdb\_reviews                 & 24800 & 200        & 25000  & accuracy   \\
piqa                          & 16013 & 100        & 1838   & accuracy   \\
samsum                        & 14732 & 200        & 819    & rougeLsum  \\
aeslc                         & 14436 & 200        & 1906   & rougeLsum  \\
bool\_q                       & 9227  & 200        & 3270   & accuracy   \\
cola                          & 8351  & 200        & 1043   & accuracy   \\
coqa                          & 7099  & 100        & 500    & nll        \\
stsb                          & 5649  & 100        & 1500   & accuracy   \\
trec                          & 5252  & 200        & 500    & accuracy   \\
wic                           & 5228  & 200        & 638    & accuracy   \\
openbookqa                    & 4957  & 200        & 500    & accuracy   \\
opinion\_abstracts            & 4790  & 200        & 1000   & rougeLsum  \\
glue\_mrpc                    & 3468  & 200        & 408    & accuracy   \\
arc                           & 2970  & 400        & 3548   & accuracy   \\
rte                           & 2290  & 200        & 277    & accuracy   \\
story\_cloze                  & 1671  & 200        & 1871   & accuracy   \\
definite\_pronoun\_resolution & 1122  & 200        & 564    & accuracy   \\
wnli                          & 605   & 30         & 71     & accuracy   \\
wsc                           & 504   & 50         & 104    & accuracy   \\
copa                          & 350   & 50         & 100    & accuracy   \\
cb                            & 200   & 50         & 56     & accuracy   \\ \bottomrule
\end{tabular}
\caption{Statistics for each dataset used in our analysis, and the primary metric used.}
\label{tab:stats}
\end{table*}

Statistics for individual datasets used in our analysis are contained in Table~\ref{tab:stats}. Following \citet{wei2022finetuned}, we cap the number of training examples per dataset at $30k$, and use up to $200$ examples for validation. We use the full test sets where available. Note that some tasks have multiple subtasks (e.g., for SQuAD~\cite{rajpurkar-etal-2016-squad} we treat SQuAD v1 and v2 as separate subtasks). We apply these limits only per \emph{subtask}, and when computing results, we take the average performance across all subtasks. The benchmark is available for download at \url{https://ct5x.s3.us-east-2.amazonaws.com/benchmark.json}. All models were trained using the T5x framework~\cite{Roberts2022ScalingUM} using the base T5 1.1 architecture with default hyper-parameters. Our pre-trained model was the \texttt{t5\_1\_1\_lm100k\_base} model released by \citet{lester-etal-2021-power}.

%% file: sections/appendix/features.tex
\section{Checkpoint selection features}
\label{app:features}
We derive a number of easy to compute features for our lightweight GBDT checkpoint selecto when evaluating transfer candidate $(t_i, t_j)$. 

\begin{itemize}[leftmargin=*]

\item \textbf{Relative performance.} We compute the relative $0$-shot and $5$-shot performance for an independently fine-tuned model (starting from the base model, trained just on $t_j$) with the sequential tuning candidate (starting from $t_i$ and training next on $t_j$). Here $k$-shot denotes the performance after $k$ updates (using a fixed batch-size). We compute the relative performance both in terms of the specific task metric (e.g., accuracy) and the token negative log-likelihood, as these can have different trends.\looseness=-1 

  \item \textbf{Weight change.} We compute the maximum and average magnitude parameter update of the checkpoint for $t_j$ relative to the base pre-trained model. This is stratified by weight group, where we differentiate between softmax, embedding, and layer parameters (the layers are collapsed into $4$ groups by layer number, e.g., layers $\{1, 2, 3\}$ are a group).
  
   \item \textbf{Update similarity.} We approximate gradient-based similarity by using the weight change of a candidate model from the base pre-trained model as an estimate of the average gradient applied (recall that all models start from the same pre-trained model). We then compare this ``average gradient'' to the weight change of the independent fine-tuning model after $5$ steps using cosine similarity. This gives an idea if the average gradient already applied to $t_i$ is similar in direction to the initial gradient computed when training directly on $t_j$. We also stratify this metric across different parameter groups (the same as for absolute weight change above).\looseness=-1
   
  \item \textbf{Task metadata.} We include binary features that indicate if the last task used to train $t_i$ is in the same manually defined family as $t_j$, and also if any task used in the full process for training $t_i$ (i.e., was used as a previous checkpoint initialization) is in the same manually defined family as $t_j$. The task families are illustrated in Figure~\ref{fig:tasks}.\looseness=-1
\end{itemize}

When comparing feature importance determined after training the GBDT, the relative $0$ and $5$-shot performance is most important, followed by gradient similarity. While this is intuitive (and also clearly helpful for our specific purpose when considering that our log-scale $\mathrm{PerAUC}$ metric heavily weights strong early performance), it is important to note that computing these features does not scale particularly well with the number of tasks, as it involves training and evaluating many models, even if only for a few updates. Gradient similarity is less costly to evaluate using our approximate method, as it only requires evaluating the weight change for all checkpoints except for the pre-trained model.

%% file: sections/appendix/additional_results.tex
\section{Additional results}
\label{app:additional_results}

We list all results per sequence type in Tables~\ref{tab:good-good} through \ref{tab:neutral-bad}, in addition to the medians in Table~\ref{tab:results}. 

\setlength{\tabcolsep}{20pt} 

  \begin{table*}[!h]
  \centering
  \begin{tabular}{l|l|lll}
  \toprule
  $\mathbf{A} \rightarrow \mathbf{C}$ & $\mathbf{B} \rightarrow \mathbf{C}$ & \multicolumn{3}{c}{$\mathbf{A} \rightarrow \mathbf{B} \rightarrow \mathbf{C}$} \\
Pairwise                        & Pairwise                        & Na\"ive & Selective & Oracle \\ \midrule
\cellcolor[HTML]{84D36B}43.31 & \cellcolor[HTML]{84D36B}22.52 & 56.97 & 43.31 & 56.97 \\
\cellcolor[HTML]{84D36B}22.52 & \cellcolor[HTML]{84D36B}47.14 & 54.55 & 47.14 & 54.55 \\
\cellcolor[HTML]{84D36B}49.27 & \cellcolor[HTML]{84D36B}50.82 & 58.54 & 58.54 & 58.54 \\
\cellcolor[HTML]{84D36B}47.14 & \cellcolor[HTML]{84D36B}50.82 & 56.69 & 56.69 & 56.69 \\
\cellcolor[HTML]{84D36B}32.59 & \cellcolor[HTML]{84D36B}32.80 & 31.97 & 32.59 & 32.80 \\
\cellcolor[HTML]{84D36B}37.56 & \cellcolor[HTML]{84D36B}56.60 & 58.50 & 56.60 & 58.50 \\
\cellcolor[HTML]{84D36B}32.59 & \cellcolor[HTML]{84D36B}59.31 & 52.37 & 52.37 & 59.31 \\
\cellcolor[HTML]{84D36B}59.31 & \cellcolor[HTML]{84D36B}32.80 & 56.58 & 59.31 & 59.31 \\
\cellcolor[HTML]{84D36B}22.71 & \cellcolor[HTML]{84D36B}30.83 & 38.42 & 22.71 & 38.42 \\
\cellcolor[HTML]{84D36B}30.58 & \cellcolor[HTML]{84D36B}28.64 & 39.13 & 39.13 & 39.13 \\
\cellcolor[HTML]{84D36B}22.71 & \cellcolor[HTML]{84D36B}30.58 & 40.50 & 40.50 & 40.50 \\
\cellcolor[HTML]{84D36B}36.50 & \cellcolor[HTML]{84D36B}30.58 & 41.57 & 36.50 & 41.57 \\
\cellcolor[HTML]{84D36B}14.07 & \cellcolor[HTML]{84D36B}12.22 & 1.37 & 12.22 & 14.07 \\
\cellcolor[HTML]{84D36B}43.37 & \cellcolor[HTML]{84D36B}14.07 & 43.77 & 43.37 & 43.77 \\
\cellcolor[HTML]{84D36B}12.22 & \cellcolor[HTML]{84D36B}43.37 & 32.90 & 12.22 & 43.37 \\
\cellcolor[HTML]{84D36B}43.37 & \cellcolor[HTML]{84D36B}17.00 & 41.58 & 41.58 & 43.37 \\
\bottomrule
\end{tabular}
\caption{Both $\mathbf{A}$ and $\mathbf{B}$ are intended to be positive.}
\label{tab:good-good}
\end{table*}

  \begin{table*}[!h]
  \centering
  \begin{tabular}{l|l|lll}
  \toprule
  $\mathbf{A} \rightarrow \mathbf{C}$ & $\mathbf{B} \rightarrow \mathbf{C}$ & \multicolumn{3}{c}{$\mathbf{A} \rightarrow \mathbf{B} \rightarrow \mathbf{C}$} \\
Pairwise                        & Pairwise                        & Na\"ive & Selective & Oracle \\ \midrule
\cellcolor[HTML]{84D36B}52.36 & \cellcolor[HTML]{FFCCC9}-310.39 & -291.80 & 52.36 & 52.36 \\
\cellcolor[HTML]{84D36B}51.16 & \cellcolor[HTML]{FFCCC9}-310.39 & -281.67 & 51.16 & 51.16 \\
\cellcolor[HTML]{84D36B}52.36 & \cellcolor[HTML]{FFCCC9}-17.99 & 49.96 & 52.36 & 52.36 \\
\cellcolor[HTML]{84D36B}51.16 & \cellcolor[HTML]{FFCCC9}-57.86 & -14.64 & 51.16 & 51.16 \\
\cellcolor[HTML]{84D36B}8.20 & \cellcolor[HTML]{FFCCC9}-16.31 & -8.35 & 8.20 & 8.20 \\
\cellcolor[HTML]{84D36B}13.66 & \cellcolor[HTML]{FFCCC9}-16.31 & 7.90 & 13.66 & 13.66 \\
\cellcolor[HTML]{84D36B}11.41 & \cellcolor[HTML]{FFCCC9}-16.31 & 10.13 & 11.41 & 11.41 \\
\cellcolor[HTML]{84D36B}14.85 & \cellcolor[HTML]{FFCCC9}-16.31 & -4.11 & 14.85 & 14.85 \\
\cellcolor[HTML]{84D36B}14.07 & \cellcolor[HTML]{FFCCC9}-17.00 & -1.44 & 0.00 & 14.07 \\
\cellcolor[HTML]{84D36B}43.37 & \cellcolor[HTML]{FFCCC9}-17.00 & 26.98 & 43.37 & 43.37 \\
\cellcolor[HTML]{84D36B}12.22 & \cellcolor[HTML]{FFCCC9}-17.00 & -3.25 & 12.22 & 12.22 \\
\cellcolor[HTML]{84D36B}17.00 & \cellcolor[HTML]{FFCCC9}-17.00 & -8.93 & 0.00 & 17.00 \\
\cellcolor[HTML]{84D36B}10.99 & \cellcolor[HTML]{FFCCC9}-18.00 & -6.10 & 10.99 & 10.99 \\
\cellcolor[HTML]{84D36B}10.99 & \cellcolor[HTML]{FFCCC9}-9.45 & 0.63 & 10.99 & 10.99 \\
\cellcolor[HTML]{84D36B}10.99 & \cellcolor[HTML]{FFCCC9}-12.91 & 1.38 & 10.99 & 10.99 \\
\cellcolor[HTML]{84D36B}10.99 & \cellcolor[HTML]{FFCCC9}-13.55 & 4.33 & 10.99 & 10.99 \\
\bottomrule
\end{tabular}
\caption{$\mathbf{A}$ is intended to be positive while $\mathbf{B}$ is intended to be negative.}
\label{tab:good-bad}
\end{table*}

  \begin{table*}[!h]
  \centering
  \begin{tabular}{l|l|lll}
  \toprule
  $\mathbf{A} \rightarrow \mathbf{C}$ & $\mathbf{B} \rightarrow \mathbf{C}$ & \multicolumn{3}{c}{$\mathbf{A} \rightarrow \mathbf{B} \rightarrow \mathbf{C}$} \\
Pairwise                        & Pairwise                        & Na\"ive & Selective & Oracle \\ \midrule
\cellcolor[HTML]{84D36B}52.20 & \cellcolor[HTML]{EFEFEF}-0.78 & 45.69 & 52.20 & 52.20 \\
\cellcolor[HTML]{84D36B}15.00 & \cellcolor[HTML]{EFEFEF}2.99 & 34.87 & 15.00 & 34.87 \\
\cellcolor[HTML]{84D36B}15.00 & \cellcolor[HTML]{EFEFEF}-3.68 & 22.51 & 15.00 & 22.51 \\
\cellcolor[HTML]{84D36B}28.70 & \cellcolor[HTML]{EFEFEF}2.99 & 23.93 & 28.70 & 28.70 \\
\cellcolor[HTML]{84D36B}17.33 & \cellcolor[HTML]{EFEFEF}-3.31 & 10.29 & 17.33 & 17.33 \\
\cellcolor[HTML]{84D36B}14.14 & \cellcolor[HTML]{EFEFEF}-3.31 & -1.38 & 14.14 & 14.14 \\
\cellcolor[HTML]{84D36B}14.14 & \cellcolor[HTML]{EFEFEF}-2.10 & 11.34 & 14.14 & 14.14 \\
\cellcolor[HTML]{84D36B}15.27 & \cellcolor[HTML]{EFEFEF}-3.31 & -0.78 & 15.27 & 15.27 \\
\cellcolor[HTML]{84D36B}59.31 & \cellcolor[HTML]{EFEFEF}0.49 & 43.87 & 59.31 & 59.31 \\
\cellcolor[HTML]{84D36B}32.80 & \cellcolor[HTML]{EFEFEF}0.49 & 27.43 & 32.80 & 32.80 \\
\cellcolor[HTML]{84D36B}59.31 & \cellcolor[HTML]{EFEFEF}5.51 & 44.56 & 59.31 & 59.31 \\
\cellcolor[HTML]{84D36B}35.85 & \cellcolor[HTML]{EFEFEF}4.09 & 32.40 & 35.85 & 35.85 \\
\cellcolor[HTML]{84D36B}-0.95 & \cellcolor[HTML]{EFEFEF}0.00 & 6.95 & -0.95 & 6.95 \\
\cellcolor[HTML]{84D36B}-0.95 & \cellcolor[HTML]{EFEFEF}4.09 & 6.95 & -0.95 & 6.95 \\
\cellcolor[HTML]{84D36B}-0.95 & \cellcolor[HTML]{EFEFEF}0.00 & -4.45 & -0.95 & 0.00 \\
\cellcolor[HTML]{84D36B}-0.95 & \cellcolor[HTML]{EFEFEF}6.95 & 6.95 & 6.95 & 6.95 \\
\bottomrule
\end{tabular}
\caption{$\mathbf{A}$ is intended to be positive while $\mathbf{B}$ is intended to be neutral.}
\label{tab:good-neutral}
\end{table*}

  \begin{table*}[!h]
  \centering
  \begin{tabular}{l|l|lll}
  \toprule
  $\mathbf{A} \rightarrow \mathbf{C}$ & $\mathbf{B} \rightarrow \mathbf{C}$ & \multicolumn{3}{c}{$\mathbf{A} \rightarrow \mathbf{B} \rightarrow \mathbf{C}$} \\
Pairwise                        & Pairwise                        & Na\"ive & Selective & Oracle \\ \midrule
\cellcolor[HTML]{FFCCC9}-16.31 & \cellcolor[HTML]{84D36B}8.20 & -12.27 & 0.00 & 8.20 \\
\cellcolor[HTML]{FFCCC9}-16.31 & \cellcolor[HTML]{84D36B}13.66 & 15.92 & 13.66 & 15.92 \\
\cellcolor[HTML]{FFCCC9}-16.31 & \cellcolor[HTML]{84D36B}14.85 & 15.13 & 14.85 & 15.13 \\
\cellcolor[HTML]{FFCCC9}-16.31 & \cellcolor[HTML]{84D36B}11.41 & 8.87 & 11.41 & 11.41 \\
\cellcolor[HTML]{FFCCC9}-12.91 & \cellcolor[HTML]{84D36B}10.99 & 1.06 & 10.99 & 10.99 \\
\cellcolor[HTML]{FFCCC9}-9.45 & \cellcolor[HTML]{84D36B}10.99 & -2.25 & 10.99 & 10.99 \\
\cellcolor[HTML]{FFCCC9}-13.55 & \cellcolor[HTML]{84D36B}10.99 & -2.80 & 10.99 & 10.99 \\
\cellcolor[HTML]{FFCCC9}-18.00 & \cellcolor[HTML]{84D36B}10.99 & -14.24 & 10.99 & 10.99 \\
\cellcolor[HTML]{FFCCC9}-17.00 & \cellcolor[HTML]{84D36B}17.00 & 5.67 & 0.00 & 17.00 \\
\cellcolor[HTML]{FFCCC9}-17.00 & \cellcolor[HTML]{84D36B}43.37 & 30.98 & 43.37 & 43.37 \\
\cellcolor[HTML]{FFCCC9}-17.00 & \cellcolor[HTML]{84D36B}12.22 & -6.83 & 12.22 & 12.22 \\
\cellcolor[HTML]{FFCCC9}-17.00 & \cellcolor[HTML]{84D36B}14.07 & 18.22 & 0.00 & 18.22 \\
\cellcolor[HTML]{FFCCC9}-310.34 & \cellcolor[HTML]{84D36B}51.16 & 39.47 & 51.16 & 51.16 \\
\cellcolor[HTML]{FFCCC9}-57.86 & \cellcolor[HTML]{84D36B}51.16 & 38.09 & 51.16 & 51.16 \\
\cellcolor[HTML]{FFCCC9}-17.99 & \cellcolor[HTML]{84D36B}52.36 & 33.78 & 33.78 & 52.36 \\
\cellcolor[HTML]{FFCCC9}-17.99 & \cellcolor[HTML]{84D36B}51.16 & 49.56 & 49.56 & 51.16 \\
\bottomrule
\end{tabular}
\caption{$\mathbf{A}$ is intended to be negative while $\mathbf{B}$ is intended to be positive.}
\label{tab:bad-good}
\end{table*}

  \begin{table*}[!h]
  \centering
  \begin{tabular}{l|l|lll}
  \toprule
  $\mathbf{A} \rightarrow \mathbf{C}$ & $\mathbf{B} \rightarrow \mathbf{C}$ & \multicolumn{3}{c}{$\mathbf{A} \rightarrow \mathbf{B} \rightarrow \mathbf{C}$} \\
Pairwise                        & Pairwise                        & Na\"ive & Selective & Oracle \\ \midrule
\cellcolor[HTML]{FFCCC9}-17.99 & \cellcolor[HTML]{FFCCC9}-310.34 & -8.89 & 0.00 & 0.00 \\
\cellcolor[HTML]{FFCCC9}-310.34 & \cellcolor[HTML]{FFCCC9}-57.86 & -66.16 & 0.00 & 0.00 \\
\cellcolor[HTML]{FFCCC9}-17.99 & \cellcolor[HTML]{FFCCC9}-310.39 & -119.44 & 0.00 & 0.00 \\
\cellcolor[HTML]{FFCCC9}-57.86 & \cellcolor[HTML]{FFCCC9}-17.99 & -49.32 & 0.00 & 0.00 \\
\cellcolor[HTML]{FFCCC9}-106.80 & \cellcolor[HTML]{FFCCC9}-36.99 & -40.34 & 0.00 & 0.00 \\
\cellcolor[HTML]{FFCCC9}-157.10 & \cellcolor[HTML]{FFCCC9}-66.13 & -372.89 & 0.00 & 0.00 \\
\cellcolor[HTML]{FFCCC9}-106.80 & \cellcolor[HTML]{FFCCC9}-5.16 & -85.35 & 0.00 & 0.00 \\
\cellcolor[HTML]{FFCCC9}-66.13 & \cellcolor[HTML]{FFCCC9}-5.16 & -22.69 & 0.00 & 0.00 \\
\cellcolor[HTML]{FFCCC9}-13.55 & \cellcolor[HTML]{FFCCC9}-18.00 & -20.60 & 0.00 & 0.00 \\
\cellcolor[HTML]{FFCCC9}-9.45 & \cellcolor[HTML]{FFCCC9}-12.91 & -7.13 & 0.00 & 0.00 \\
\cellcolor[HTML]{FFCCC9}-9.45 & \cellcolor[HTML]{FFCCC9}-18.00 & -5.68 & 0.00 & 0.00 \\
\cellcolor[HTML]{FFCCC9}-12.91 & \cellcolor[HTML]{FFCCC9}-18.00 & -5.48 & 0.00 & 0.00 \\
\cellcolor[HTML]{FFCCC9}-16.89 & \cellcolor[HTML]{FFCCC9}-21.00 & -18.95 & 0.00 & 0.00 \\
\cellcolor[HTML]{FFCCC9}-27.96 & \cellcolor[HTML]{FFCCC9}-16.89 & -22.73 & 0.00 & 0.00 \\
\cellcolor[HTML]{FFCCC9}-27.96 & \cellcolor[HTML]{FFCCC9}-21.00 & -26.68 & 0.00 & 0.00 \\
\cellcolor[HTML]{FFCCC9}-21.00 & \cellcolor[HTML]{FFCCC9}-27.96 & -32.79 & 0.00 & 0.00 \\
\bottomrule
\end{tabular}
\caption{}
\caption{Both $\mathbf{A}$ and $\mathbf{B}$ are intended to be negative.}
\end{table*}

  \begin{table*}[!h]
  \centering
  \begin{tabular}{l|l|lll}
  \toprule
  $\mathbf{A} \rightarrow \mathbf{C}$ & $\mathbf{B} \rightarrow \mathbf{C}$ & \multicolumn{3}{c}{$\mathbf{A} \rightarrow \mathbf{B} \rightarrow \mathbf{C}$} \\
Pairwise                        & Pairwise                        & Na\"ive & Selective & Oracle \\ \midrule
\cellcolor[HTML]{FFCCC9}-17.00 & \cellcolor[HTML]{EFEFEF}-12.78 & -17.39 & -12.78 & 0.00 \\
\cellcolor[HTML]{FFCCC9}-17.00 & \cellcolor[HTML]{EFEFEF}-20.57 & -16.32 & -20.57 & 0.00 \\
\cellcolor[HTML]{FFCCC9}-17.00 & \cellcolor[HTML]{EFEFEF}-2.10 & -22.33 & 0.00 & 0.00 \\
\cellcolor[HTML]{FFCCC9}-17.00 & \cellcolor[HTML]{EFEFEF}2.22 & 7.99 & 0.00 & 7.99 \\
\cellcolor[HTML]{FFCCC9}-106.80 & \cellcolor[HTML]{EFEFEF}5.21 & 1.10 & 5.21 & 5.21 \\
\cellcolor[HTML]{FFCCC9}-77.67 & \cellcolor[HTML]{EFEFEF}-0.73 & -10.93 & 0.00 & 0.00 \\
\cellcolor[HTML]{FFCCC9}-77.67 & \cellcolor[HTML]{EFEFEF}8.12 & 1.48 & 0.00 & 8.12 \\
\cellcolor[HTML]{FFCCC9}-66.13 & \cellcolor[HTML]{EFEFEF}-0.73 & -3.65 & 0.00 & 0.00 \\
\cellcolor[HTML]{FFCCC9}-16.89 & \cellcolor[HTML]{EFEFEF}-2.96 & -14.71 & 0.00 & 0.00 \\
\cellcolor[HTML]{FFCCC9}-27.96 & \cellcolor[HTML]{EFEFEF}-3.43 & -15.41 & 0.00 & 0.00 \\
\cellcolor[HTML]{FFCCC9}-21.00 & \cellcolor[HTML]{EFEFEF}-4.53 & -22.14 & 0.00 & 0.00 \\
\cellcolor[HTML]{FFCCC9}-27.96 & \cellcolor[HTML]{EFEFEF}-4.53 & -14.35 & 0.00 & 0.00 \\
\cellcolor[HTML]{FFCCC9}-6.33 & \cellcolor[HTML]{EFEFEF}-2.75 & -2.77 & 0.00 & 0.00 \\
\cellcolor[HTML]{FFCCC9}-6.33 & \cellcolor[HTML]{EFEFEF}2.81 & -5.74 & 0.00 & 2.81 \\
\cellcolor[HTML]{FFCCC9}-6.33 & \cellcolor[HTML]{EFEFEF}-0.97 & -4.85 & 0.00 & 0.00 \\
\cellcolor[HTML]{FFCCC9}-6.33 & \cellcolor[HTML]{EFEFEF}-1.21 & -3.65 & 0.00 & 0.00 \\
\bottomrule
\end{tabular}
\caption{$\mathbf{A}$ is intended to be negative while $\mathbf{B}$ is intended to be neutral.}
\label{tab:bad-neutral}
\end{table*}

  \begin{table*}[!h]
  \centering
  \begin{tabular}{l|l|lll}
  \toprule
  $\mathbf{A} \rightarrow \mathbf{C}$ & $\mathbf{B} \rightarrow \mathbf{C}$ & \multicolumn{3}{c}{$\mathbf{A} \rightarrow \mathbf{B} \rightarrow \mathbf{C}$} \\
Pairwise                        & Pairwise                        & Na\"ive & Selective & Oracle \\ \midrule
\cellcolor[HTML]{EFEFEF}-0.70 & \cellcolor[HTML]{84D36B}11.41 & 8.71 & 11.41 & 11.41 \\
\cellcolor[HTML]{EFEFEF}-12.99 & \cellcolor[HTML]{84D36B}13.66 & 10.29 & 13.66 & 13.66 \\
\cellcolor[HTML]{EFEFEF}2.67 & \cellcolor[HTML]{84D36B}13.66 & 10.06 & 13.66 & 13.66 \\
\cellcolor[HTML]{EFEFEF}-0.70 & \cellcolor[HTML]{84D36B}8.20 & 6.25 & 8.20 & 8.20 \\
\cellcolor[HTML]{EFEFEF}0.16 & \cellcolor[HTML]{84D36B}16.38 & 8.50 & 16.38 & 16.38 \\
\cellcolor[HTML]{EFEFEF}-7.90 & \cellcolor[HTML]{84D36B}7.83 & -0.79 & -7.90 & 7.83 \\
\cellcolor[HTML]{EFEFEF}0.16 & \cellcolor[HTML]{84D36B}10.45 & 11.52 & 10.45 & 11.52 \\
\cellcolor[HTML]{EFEFEF}2.71 & \cellcolor[HTML]{84D36B}10.45 & 1.90 & 1.90 & 10.45 \\
\cellcolor[HTML]{EFEFEF}-40.79 & \cellcolor[HTML]{84D36B}52.36 & 51.84 & 51.84 & 52.36 \\
\cellcolor[HTML]{EFEFEF}27.80 & \cellcolor[HTML]{84D36B}52.36 & 44.12 & 52.36 & 52.36 \\
\cellcolor[HTML]{EFEFEF}-40.79 & \cellcolor[HTML]{84D36B}51.16 & 48.45 & 51.16 & 51.16 \\
\cellcolor[HTML]{EFEFEF}27.80 & \cellcolor[HTML]{84D36B}51.16 & 47.42 & 51.16 & 51.16 \\
\cellcolor[HTML]{EFEFEF}13.47 & \cellcolor[HTML]{84D36B}34.11 & 36.22 & 34.11 & 36.22 \\
\cellcolor[HTML]{EFEFEF}-6.09 & \cellcolor[HTML]{84D36B}45.62 & 54.49 & 45.62 & 54.49 \\
\cellcolor[HTML]{EFEFEF}-9.01 & \cellcolor[HTML]{84D36B}80.13 & 76.83 & 80.13 & 80.13 \\
\cellcolor[HTML]{EFEFEF}-13.75 & \cellcolor[HTML]{84D36B}57.81 & 28.08 & 0.00 & 57.81 \\
\bottomrule
\end{tabular}
\caption{}
\caption{$\mathbf{A}$ is intended to be neutral while $\mathbf{B}$ is intended to be positive.}
\end{table*}

  \begin{table*}[!h]
  \centering
  \begin{tabular}{l|l|lll}
  \toprule
  $\mathbf{A} \rightarrow \mathbf{C}$ & $\mathbf{B} \rightarrow \mathbf{C}$ & \multicolumn{3}{c}{$\mathbf{A} \rightarrow \mathbf{B} \rightarrow \mathbf{C}$} \\
Pairwise                        & Pairwise                        & Na\"ive & Selective & Oracle \\ \midrule
\cellcolor[HTML]{EFEFEF}2.81 & \cellcolor[HTML]{FFCCC9}-6.33 & -1.21 & 0.00 & 2.81 \\
\cellcolor[HTML]{EFEFEF}-0.97 & \cellcolor[HTML]{FFCCC9}-6.33 & -3.13 & -0.97 & 0.00 \\
\cellcolor[HTML]{EFEFEF}-3.74 & \cellcolor[HTML]{FFCCC9}-6.33 & -4.13 & 0.00 & 0.00 \\
\cellcolor[HTML]{EFEFEF}-3.64 & \cellcolor[HTML]{FFCCC9}-6.33 & -2.45 & 0.00 & 0.00 \\
\cellcolor[HTML]{EFEFEF}-2.96 & \cellcolor[HTML]{FFCCC9}-21.00 & -22.89 & 0.00 & 0.00 \\
\cellcolor[HTML]{EFEFEF}-2.96 & \cellcolor[HTML]{FFCCC9}-16.89 & -12.74 & 0.00 & 0.00 \\
\cellcolor[HTML]{EFEFEF}-4.53 & \cellcolor[HTML]{FFCCC9}-16.89 & -14.76 & 0.00 & 0.00 \\
\cellcolor[HTML]{EFEFEF}-3.32 & \cellcolor[HTML]{FFCCC9}-16.89 & -17.37 & 0.00 & 0.00 \\
\cellcolor[HTML]{EFEFEF}8.12 & \cellcolor[HTML]{FFCCC9}-5.16 & 1.27 & 0.00 & 8.12 \\
\cellcolor[HTML]{EFEFEF}8.12 & \cellcolor[HTML]{FFCCC9}-5.75 & 3.54 & 0.00 & 8.12 \\
\cellcolor[HTML]{EFEFEF}5.21 & \cellcolor[HTML]{FFCCC9}-66.13 & -9.05 & 5.21 & 5.21 \\
\cellcolor[HTML]{EFEFEF}5.21 & \cellcolor[HTML]{FFCCC9}-36.99 & -33.14 & 5.21 & 5.21 \\
\cellcolor[HTML]{EFEFEF}-1.66 & \cellcolor[HTML]{FFCCC9}-17.00 & -12.37 & -1.66 & 0.00 \\
\cellcolor[HTML]{EFEFEF}-20.57 & \cellcolor[HTML]{FFCCC9}-17.00 & -25.37 & -20.57 & 0.00 \\
\cellcolor[HTML]{EFEFEF}-2.10 & \cellcolor[HTML]{FFCCC9}-17.00 & -14.88 & 0.00 & 0.00 \\
\cellcolor[HTML]{EFEFEF}1.28 & \cellcolor[HTML]{FFCCC9}-17.00 & -21.27 & 0.00 & 1.28 \\
\bottomrule
\end{tabular}
\caption{$\mathbf{A}$ is intended to be neutral while $\mathbf{B}$ is intended to be negative.}
\label{tab:neutral-bad}
\end{table*}